\pdfoutput=1

\documentclass[11pt]{article}

\usepackage[final]{acl}

\usepackage{times}
\usepackage{latexsym}

\usepackage[T1]{fontenc}

\usepackage[utf8]{inputenc}

\usepackage{microtype}

\usepackage{inconsolata}

\usepackage{graphicx}

\usepackage{amsmath}
\usepackage{amssymb}
\usepackage{mathtools}
\usepackage{amsthm}
\usepackage{pgfplots}
\usepackage[capitalise]{cleveref} 
\usepackage{enumitem} 
\usepackage{algorithm} 
\usepackage{algorithmic} 
\usepackage{booktabs} 
\usepackage{colortbl} 
\usepackage{multirow} 
\definecolor{mygray}{gray}{0.9} 
\usepackage{subfigure} 
\theoremstyle{plain}
\newtheorem{theorem}{Theorem}

\theoremstyle{definition}

\theoremstyle{remark}
\newtheorem{remark}{Remark}
\newcommand{\TM}{\mathbf{M}_{i}}
\newcommand{\TP}{\mathbf{P}_{i}}
\newcommand{\MM}{\mathbf{M}}
\newcommand{\bfu}{\mathbf{u}}
\newcommand{\bfv}{\mathbf{v}}
\newcommand{\bfb}{\mathbf{b}}
\newcommand{\Real}{\mathbb{R}}
\newcommand{\SigM}{\mathbf{\Sigma}}
\newcommand{\UM}{\mathbf{U}}
\newcommand{\VM}{\mathbf{V}}
\newcommand{\method}{STF}

\usepackage{todo}
\title{Superpose Task-specific Features for Model Merging}

\author{
\textbf{Haiquan~Qiu\textsuperscript{1}},
\textbf{You~Wu\textsuperscript{1}},
\textbf{Dong~Li\textsuperscript{2}},
\textbf{Jianmin~Guo\textsuperscript{2}},
\textbf{Quanming~Yao\textsuperscript{1,3$\ast$}}, \\
\textsuperscript{1}Tsinghua University,
\textsuperscript{2}Huawei, \\
\textsuperscript{3}State Key laboratory of Space Network and Communications
\\
\small{
\textbf{Correspondence:} \href{mailto:qyaoaa@tsinghua.edu.cn}{qyaoaa@tsinghua.edu.cn}
}
}

\begin{document}
\maketitle
\begin{abstract}
Model merging enables powerful capabilities in neural networks without requiring additional training. In this paper, we introduce a novel perspective on model merging by leveraging the fundamental mechanisms of neural network representation. Our approach is motivated by the linear representation hypothesis, which states that neural networks encode information through linear combinations of feature vectors. We propose a method that superposes task-specific features from individual models into a merged model. Our approach specifically targets linear transformation matrices, which are crucial for feature activation and extraction in deep networks. By formulating the merging process as a linear system, we can preserve task-specific features from individual models and create merged models that effectively maintain multi-task capabilities compared to existing methods. Extensive experiments across diverse benchmarks and models demonstrate that our method outperforms existing techniques. Code is available at \url{https://github.com/LARS-research/STF}.
\end{abstract}



\section{Introduction}

\begin{figure}[h]
    \centering
    \subfigure[Task-specific feature preservation of merged methods]{
        \includegraphics[width=1.0\columnwidth]{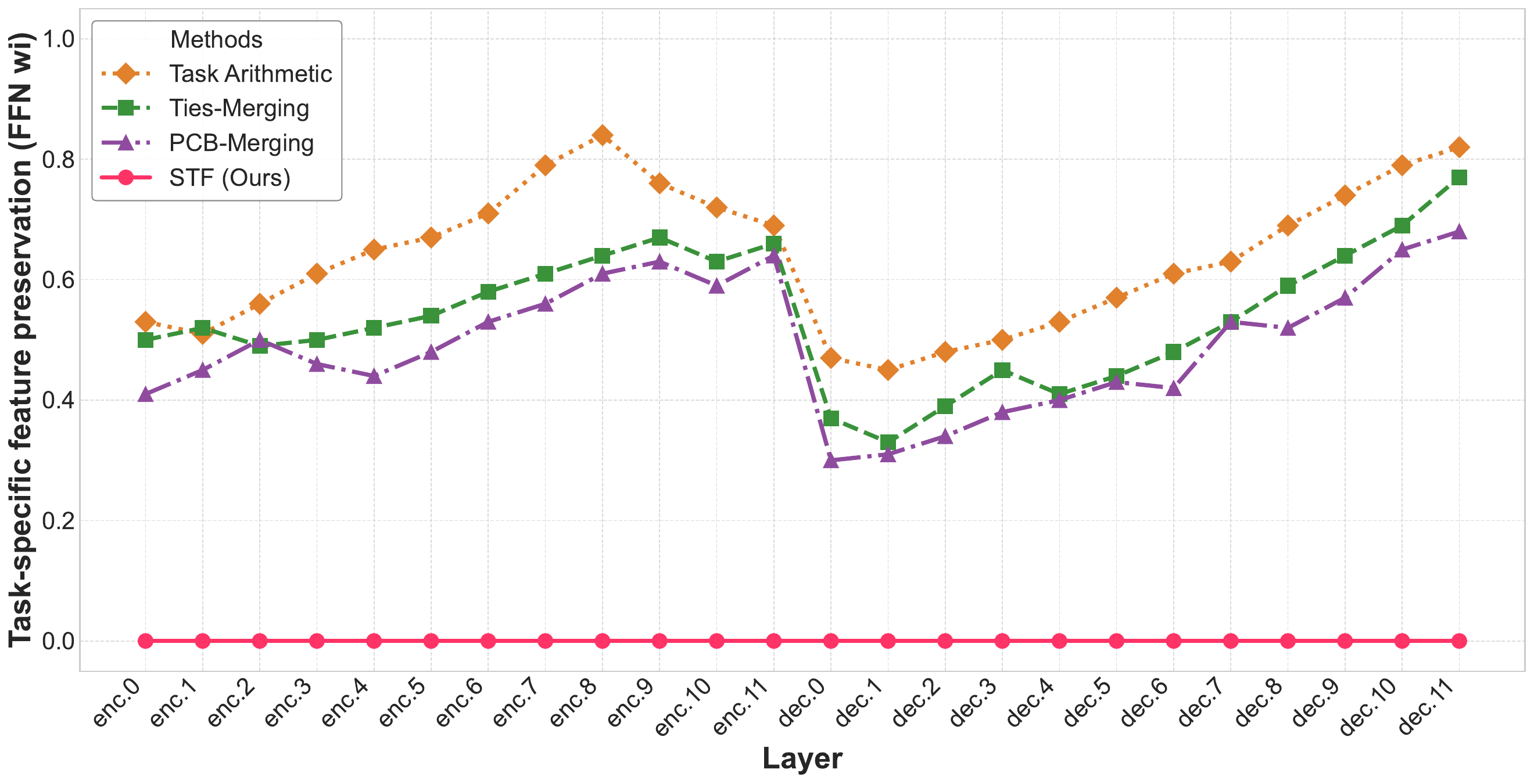}
        \label{fig:abs_direction_mismatch_ffn_wi_overall_avg}
    }
    \\
    \subfigure[Performance on T5 model merging]{
        \includegraphics[width=1.0\columnwidth]{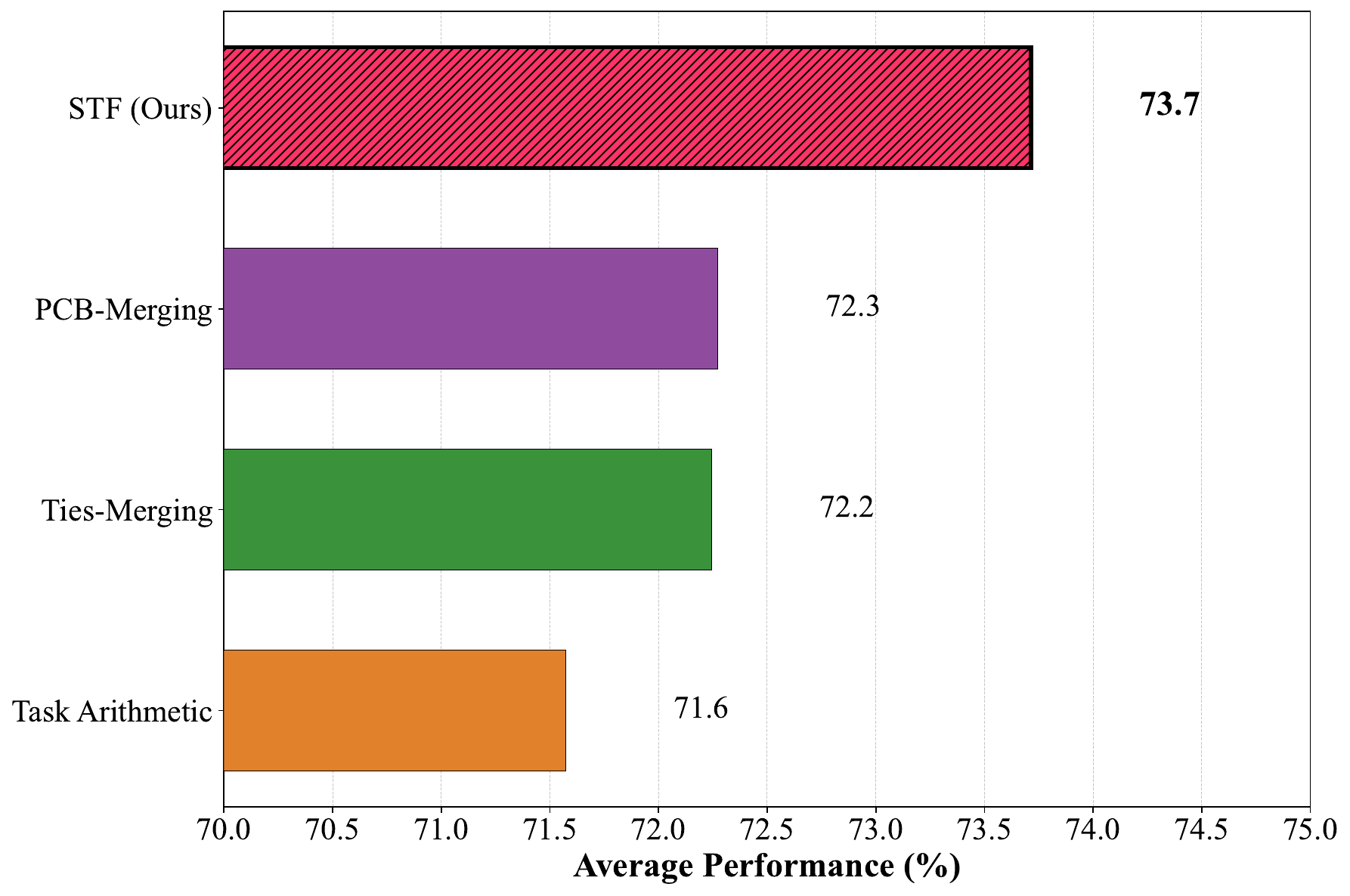}
        \label{fig:t5_average_performance_barchart}
    }
    \caption{We measure the task-specific feature preservation of merged methods in (a) (refer to Appendix \ref{app:feature_preservation}), where smaller values indicate better preservation. The merging performance in (b) strongly correlates with task-specific feature preservation.}
    \label{fig:introduce}
\end{figure}

Because of the increasing resource demands of training models, model merging has emerged as an efficient approach to consolidate capabilities from specialist models while reducing storage and deployment costs. Various model merging methods have been proposed, including parameter averaging \cite{choshen2022fusing,wortsman2022model} and task vectors \cite{ilharco2022editing,du2024parameter}. However, these methods primarily focus on parameter-level operations and do not explicitly incorporate the fundamental working mechanisms of neural networks in their design.

We argue that a principled approach to model merging should be conditioned on 
how deep neural networks represent and process information. Therefore, to design our merging method, 
we draw upon the linear representation hypothesis~\cite{mikolov2013linguistic,arora2016latent,olah2020zoom}, 
which states that neural network representations 
can be decomposed into combinations of feature vectors. 
Recent works in mechanistic interpretability~\cite{elhage2022toy,bricken2023monosemanticity,templeton2024scaling,lindsey2025biology} validate the hypothesis and also reveal that these representations often contain features both related and unrelated to the model input. This phenomenon motivates our approach: linearly superposing features from individual models into the representation of the merged model can preserve task-specific capabilities.

Therefore, we propose a method that Superposes Task-specific Features(STF) from individual finetuned models. Our approach specifically targets linear transformation matrices, which comprise the majority of model parameters and form the foundation of the linear representation hypothesis. These matrices activate features through inner products with row vectors and extract new features through linear combinations via column vectors, making them crucial for detecting and processing information in deep neural networks. We design the merged linear transformation matrices to preserve the output features from individual models when processing the same input. By identifying task-specific features through singular value decomposition of task matrices, we formulate feature superposition as a linear system that derives the optimal merged task matrix. The final merged model is created by adding this merged matrix to the pre-trained model. Experiments verified the strong positive correlation between feature preservation and performance improvement(see \cref{fig:introduce}). We validate our method on multiple benchmark datasets and models. The results demonstrate that our method consistently outperforms existing model merging techniques.

\paragraph{Notations} Throughout this paper, we use bold uppercase letters (e.g., $\mathbf{X}$) to denote matrices, bold lowercase letters (e.g., $\mathbf{x}$) to denote vectors, and regular lowercase letters (e.g., $x$) to denote scalars. For a matrix $\mathbf{X}$, its transpose is denoted as $\mathbf{X}^\top$. We denote task $i$ as $\mathsf{T}_i$ and its corresponding model parameters as $\theta_i$. For linear transformations, a linear transformation matrix in the model trained on task $i$ is represented as $\TP$. The subscript $_{\text{pre}}$ indicates pre-trained parameters or matrices - for example, $\theta_{\text{pre}}$ and $\mathbf{P}_{\text{pre}}$ denote the parameters and linear transformation matrices of the pre-trained model, respectively. The notation $\mathbf{X} \circ \mathbf{Y}$ denotes the Hadamard product of matrices $\mathbf{X}$ and $\mathbf{Y}$.

\section{Related Work}


\subsection{Model Merging of Fine-tuned Models}
Model merging is a technique that combines multiple models into a single model to enhance performance or enable the model to perform multiple tasks. 
Previous studies have shown that averaging the weights of multiple models fine-tuned from the same pre-trained initialization is a promising approach for model merging. 
Fisher Merging~\cite{matena2022merging} advances beyond simple averaging by utilizing the Fisher information matrix to assess the importance of individual parameters, which are then weighted accordingly during the merging process. Similarly, RegMean~\cite{jin2022dataless} forms a linear regression problem with extra data for each layer and offers a closed-form solution for the merged model's parameters by solving the regression problem.

Beyond parameter averaging, Task Arithmetic~\cite{ilharco2022editing} introduces task vectors and adding the task vectors of individual tasks to merge model, demonstrating their effectiveness and lightweight nature in facilitating cross-task generalization. Building on this concept, PEM Composition~\cite{zhang2023composing} extends the task arithmetic framework to merge LoRA~\cite{hu2021lora}, while Ties-Merging~\cite{yadav2023ties} addresses task conflicts by resetting redundant parameters and resolving sign conflicts. These methods, however, use a single merging coefficient across all task vectors, which limits their flexibility. In contrast, Lorahub~\cite{huang2023lorahub} and AdaMerging~\cite{yang2023adamerging} use different coefficients for enhanced adaptability. Lorahub's performance is limited as it only searches for coefficients at the task level, while AdaMerging requires complex training and unlabeled test datasets, making it applicable solely to classification problems. DARE~\cite{yu2024language} proposes drop and rescale as preprocessing steps when merging fine-tuned LLMs. PCB-Merging~\cite{du2024parameter} is a lightweight, training-free technique for model merging that balances parameter competition by intra-balancing parameter significance within tasks and inter-balancing parameter similarities across tasks, effectively enhancing performance across various scenarios.


\begin{figure*}[t]
    \centering
    \includegraphics[width=0.99\textwidth]{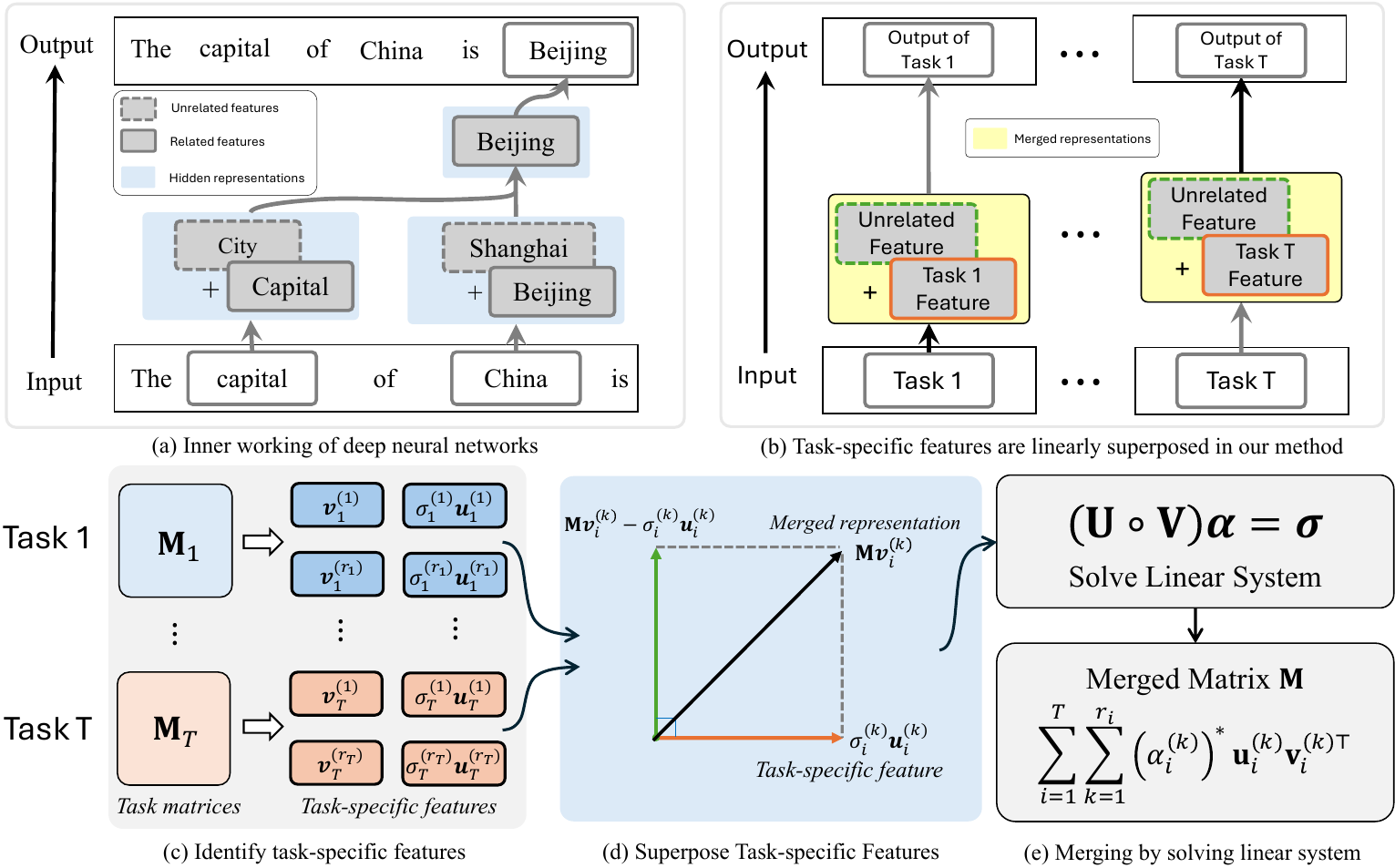}
    \caption{Deep neural networks (a) linearly add both relevant and irrelevant features in hidden representations. This insight motivates our approach to model merging through linear superposition of task-specific features. The second row illustrates how {\method} preserves and combines essential features from individual models to create an effective merged method.}\label{fig:merge-overview}
\end{figure*}
\subsection{Linear Representation Hypothesis}
The linear representation hypothesis states that neural networks encode information by summing up feature vectors~\cite{mikolov2013linguistic,arora2016latent,olah2020zoom}, i.e., a layer of a network represents a set of features as a weighted sum of 
task-associated vectors.
This hypothesis has been observed in various models, including word embeddings~\cite{mikolov2013linguistic,conneau2017word}, sentence embeddings~\cite{bowman2015generating}, Transformer language models~\cite{meng2022locating,hendel2023context}, and vision-language models~\cite{trager2023linear,perera2023prompt}.
The hypothesis has been exploited in various fields, especially in probing~\cite{alain2018understanding, belinkov2022probing} and interpretability~\cite{nostalgebraist2020logitlens,elhage2022toy,bricken2023monosemanticity,gao2024scaling,lindsey2025biology}.

The linear representation hypothesis has been validated in recent research on mechanistic interpretability of language models~\cite{elhage2022toy,bricken2023monosemanticity,templeton2024scaling}, which showed that neural networks learn to represent meaningful features through linear combinations of neurons.

\section{Method}


\subsection{Problem Formulation}
\label{ssec:setup_and_motivation}

We start with a set of tasks \(\{\mathsf{T}_{1}, \ldots, \mathsf{T}_{T}\}\) and various pre-trained models. 
The objective is to fine-tune these models either by updating all parameters or using 
parameter-efficient fine-tuning (PEFT) methods. 
The goal of model merging is to combine multiple fine-tuned models into a single model that can perform all tasks \(\{\mathsf{T}_{1}, \ldots, \mathsf{T}_{T}\}\) effectively without requiring access to training data.

To superpose task-specific features, we focus on merging the linear transformation matrices of the models, as these matrices are core components contributed to the linear representation hypothesis~\cite{elhage2022toy}. In this paper, instead of merging linear transformation matrices directly, we merge the \textit{task matrices} $\TM = \TP - \mathbf{P}_{\text{pre}} \in \mathbb{R}^{m\times n}$ where $\TP$ is a linear transformation matrix for model of task $\mathsf{T}_i$, $\mathbf{P}_{\text{pre}}$ is the linear transformation matrix for the pre-trained model, as the task matrices contain the task-specific information compared to model parameters that contains both task-specific and pre-trained information.
Specifically, we merge task matrices $\{ \TM \}$, i.e., into a single merged matrix $\MM \in \mathbb{R}^{m\times n}$.
For an input feature $\mathbf{x}\in\mathbb{R}^n$, we aim to ensure that the output of the merged model $\MM \mathbf{x}$ maintains the features of the outputs $\TM \mathbf{x}$ of the models for task $\mathsf{T}_i$.
The final merged model is then constructed by adding the merged matrix $\MM$ to the pre-trained model with a scaling factor $\gamma$: $\mathbf{P} = \mathbf{P}_{\text{pre}} + \gamma \MM$.

\subsection{Task-specific Feature Superposition}\label{ssec:feature-superposition}

\paragraph{Identify Task-specific Features}
To merge the linear transformation matrices to linearly superpose task-specific features, we need to identify these features first.  In this paper, we choose the singular vectors of the task matrices as the task-specific features, i.e., decomposing $\TM$ as $\TM=\sum_{k=1}^{r_i} \sigma_{i}^{(k)} \bfu_{i}^{(k)} \bfv_{i}^{(k)\top}$, where $\sigma_{i}^{(k)}$ is the $k$-th singular value, and $\bfu_{i}^{(k)}\in\mathbb{R}^{m}$ and $\bfv_{i}^{(k)}\in\mathbb{R}^{n}$ are the corresponding left and right singular vectors. This decomposition naturally provides task-specific features: the right singular vectors $\bfv_{i}^{(k)}$ form input features for the input $\mathbf{x}$, and $\sigma_{i}^{(k)} \bfu_{i}^{(k)}$ forms the task-specific feature for the output of task matrices.

\begin{remark}
  \it
  Singular vectors also reveal how the model processes information, i.e., the right singular vectors act as feature detectors through inner products with inputs, while the left singular vectors and singular values determine how detected features are combined into outputs.
  Refer to \cref{app:svd} for detailed analysis of singular vector. 
\end{remark}


\paragraph{Objective of Superposing Features}
Given the input features $\bfv_{i}^{(k)}$ of task $i$, we want that after the features transformed by merged matrix $\MM$, the output $\MM \bfv_{i}^{(k)}$ contains the task-specific output feature $\sigma_{i}^{(k)} \bfu_{i}^{(k)}$. This leads to the following objective:
\begin{equation}\label{eq:obj}
    \big<\sigma_i^{(k)} \bfu_{i}^{(k)}, \underbrace{\MM \bfv_{i}^{(k)} - \sigma_i^{(k)} \bfu_i^{(k)}}_{\text{Unrelated features}} \big> = 0,
\end{equation}
where $\left<\cdot,\cdot\right>$ is the inner product.
In \eqref{eq:obj}, we first obtain the unrelated features by subtracting the task-specific output feature $\sigma_{i}^{(k)} \bfu_{i}^{(k)}$ from the transformed feature $\MM \bfv_{i}^{(k)}$. The inner product \eqref{eq:obj} should be zero, which means that the unrelated features are orthogonal to the task-specific output feature. This ensures that the merged matrix $\MM$ preserves the task-specific feature $\sigma_{i}^{(k)} \bfu_{i}^{(k)}$ while allowing other features to be superposed linearly.

\begin{remark}[Comparison against RegMean]
\it Ideally, we would fully preserve features by requiring $\MM \bfv_{i}^{(k)} = \TM \bfv_{i}^{(k)}$ for all tasks (similar to RegMean). However, due to the vast number of features, merged parameters cannot simultaneously maintain all features, as this would lead to an overdetermined system of equations. Instead, our approach, {\method}, superposes task-specific features rather than requiring exact feature preservation, resulting in increased representation efficiency and superior performance.
\end{remark}

\subsection{Merging by Solving Linear System}
\label{ssec:method}

To merge the task matrices, we can see that the superposition objective~\eqref{eq:obj} is a linear equation with the unknowns being the merging matrices $\MM$. However, direct solving $\MM$ is not feasible because of numerous unknowns and equations in the linear system consisting of \eqref{eq:obj}.
For efficient merging, we instead merge the task matrices $\TM$ by decomposing them via SVD and merging their singular decomposition components:
\begin{equation}\label{eq:merge}
\MM = 
\sum\nolimits_{i=1}^T 
\sum\nolimits_{k=1}^{r_{i}} \alpha_{i}^{(k)} \bfu_{i}^{(k)} \bfv_{i}^{(k) \top},
\end{equation}
where $\alpha_{i}^{(k)}$ are the merging weights. This reduces the number of unknowns and equations to $r=\sum_{i=1}^{T} r_{i}$, where $r_i$ is the rank of task matrix $\TM$. By merging in the singular space, we can ensure that the resulting problem becomes a linear system of equations with the following theorem (proved in Appendix~\ref{app:linear}).
\begin{theorem}
\label{thm:linear}
Given the task matrices $\TM$ with their SVD decompositions, the merging weights $\alpha_{i}^{(k)}$ can be obtained by solving the linear system:
\begin{equation}\label{eq:linear-system}
(\UM \circ \VM) \boldsymbol{\alpha} = \boldsymbol{\sigma},
\end{equation}
where 
\begin{align*}
  \UM &= \begin{bmatrix}
    \bfu_{1}^{(1)\top}\!\bfu_{1}^{(1)}\! & \!\!\! \cdots \!\!\! & \bfu_{1}^{(1)\top}\! \bfu_{T}^{(r_{T})}\! \\
    \vdots & \ddots & \vdots \\
    \bfu_{T}^{(r_{T})\top}\! \bfu_{1}^{(1)}\!& \cdots & \bfu_{T}^{(r_{T})\top}\! \bfu_{T}^{(r_{T})}
\end{bmatrix}, \\
\VM &= \begin{bmatrix}
    \bfv_{1}^{(1) \top}\! \bfv_{1}^{(1)} & \!\!\! \cdots \!\!\! & \bfv_{T}^{(r_{T}) \top}\! \bfv_{1}^{(1)} \\
    \vdots & \ddots & \vdots \\
    \bfv_{1}^{(1) \top}\! \bfv_{T}^{(r_{T})} & \cdots & \bfv_{T}^{(r_{T}) \top}\! \bfv_{T}^{(r_{T})}
\end{bmatrix}, \\
\boldsymbol{\alpha} &= 
[
    \alpha_{1}^{(1)},
    \cdots,
    \alpha_{T}^{(r_{T})}
] \;\;\text{and} \;\;
\boldsymbol{\sigma} =       [ \sigma_{1}^{(1)},
    \cdots,
    \sigma_{T}^{(r_{T})}
].
\end{align*}
\end{theorem}

By solving the linear system \eqref{eq:linear-system}, we obtain the optimal weights $\boldsymbol{\alpha}^*$ for merging task matrices according to \eqref{eq:merge}. 
The merged linear transformation matrix $\mathbf{P}$ is then obtained by adding the merged task matrix $\MM$ to the pre-trained matrix $\mathbf{P}_{\text{pre}}$ with a scaling factor $\gamma$: $\mathbf{P} = \mathbf{P}_{\text{pre}} + \gamma \MM$. 

\begin{algorithm}[t]
	\caption{Merging task matrices: $\mathsf{{\method}}(\{\TM\})$}
	\label{alg:ssf}
  \small
	\begin{algorithmic}
		\STATE {\bfseries Input:} task matrices $\TM,i=1,\ldots,T$;
		\STATE Apply SVD to $\TM$ to obtain $\sigma_{i}^{(k)}$, $\bfu_{i}^{(k)}$, and $\bfv_{i}^{(k)}$;
		\STATE Prepare $\UM$, and $\VM$;
		\STATE Solve $(\UM \circ \VM) \boldsymbol{\alpha} = \boldsymbol{\sigma}$ to obtain $\boldsymbol{\alpha}^*$;
		\STATE Obtain the merged task matrix $\MM$ by \eqref{eq:merge};
		\STATE {\bfseries Return:} merged task matrices $\MM$.
	\end{algorithmic}
\end{algorithm}

The algorithm for merging task matrices is summarized in \cref{alg:ssf} and \cref{fig:merge-overview}. 
Because there are $T$ tasks matrices of size $m \times n$, SVD of these matrices takes the time complexity $O(T m n^2)$. To merge $T$ task matrices, the linear system has $r$ equations and variables to solve, which takes the time complexity $O(r^3)$. Therefore, the overall time complexity of {\method} for merging $T$ task matrices is $O(T m n^2 + r^3)$. See \cref{ssec:additional_results} for the time consumption of model merging with {\method} for various models.

\subsection{Complete Algorithm}


Except the linear transformation matrices, there are parameters in the model that need to be merged, such as biases, normalization parameters, and embeddings. 
For biases and embeddings, we merge them with task arithmetic, i.e., adding their task vectors together. For normalization parameters, we merge them by averaging the normalization parameters of the individual tasks because the normalization can be seen as a linear transformation with diagonal matrix.

We also follow Ties-Merging~\cite{yadav2023ties} and apply a trimming step. This involves keeping only the top $\eta\%$ parameters by magnitude while setting the remaining parameters to zero. This preprocessing step helps reduce noise and focus on the most significant parameters during merging. We present the complete algorithm for merging models in \cref{alg:ssf-complete}.

\begin{algorithm}[tb]
	\caption{Complete Merging Algorithm}
	\label{alg:ssf-complete}
	\begin{algorithmic}[1]
		\STATE {\bfseries Input:} fine-tuned models $\{\theta_1,\ldots,\theta_T\}$, pre-trained model $\theta_{\text{pre}}$
		\STATE {\bfseries Parameters:} sparsity ratio $\eta$, scaling factor $\gamma$
		\FOR{each linear transformation layer}
		\STATE Extract task matrices $\TM = \TP - \mathbf{P}_{\text{pre}}$ for $i=1,\ldots,T$
		\STATE Keep top $\eta$ parameters by magnitude in each $\TM$
		\STATE $\MM \leftarrow \mathsf{{\method}}(\{\TM\}_{i = 1}^T)$ \COMMENT{Algorithm \ref{alg:ssf}}
		\STATE Merged matrix $\mathbf{P} = \mathbf{P}_{\text{pre}} + \gamma \MM$
		\ENDFOR
		\FOR{each bias, embedding, normalization layer}
		\STATE Extract task vectors $\tau_i = \theta_i - \theta_{\text{pre}}$ for $i=1,\ldots,T$
		\STATE Keep top $\eta$ parameters by magnitude in each $\tau_i$
		\IF{normalization layer}
		\STATE Merged parameters $\leftarrow \mathsf{mean}(\{\tau_1,\ldots,\tau_T\})$
		\ELSE
		\STATE Merged parameters $\leftarrow \gamma \sum_{i=1}^T \tau_i$
		\ENDIF
		\ENDFOR
		\STATE {\bfseries Return:} merged model parameters
	\end{algorithmic}
\end{algorithm}

\vspace{-0.2cm}
\subsection{Discussion}\label{sec:discussion}

\paragraph{Merging Task Matrices or Fine-tuned Matrices?}
In this paper, our focus is on merging task matrices instead of fine-tuning matrices. We have discovered that fine-tuned linear transformation matrices tend to have more shared features across tasks.
Therefore, merging the fine-tuned linear transformation matrix is not as efficient as merging task matrix because certain overlapping direction of singular vectors correspond to these common features. On the contrary, the singular vectors of the task matrix contain features that are more specific to individual tasks, making it a more effective approach for merging. See the results of merging fine-tuned matrices in \cref{ssec:additional_results}.

\paragraph{Feature Interference and Scaling Factor}
In {\method}, we scale the merged matrix $\MM$ to reduce interference between task-specific features. During merging, features from different tasks can interfere since we combine them through linear superposition. While using singular vectors as basis helps minimize interference within each task, some features may still be diminished when their directions conflict with features from other tasks. The scaling factor $\gamma$ plays a crucial role in managing this tradeoff - a small $\gamma$ reduces interference but may weaken task-specific features, while a large $\gamma$ better preserves individual task capabilities but risks amplifying interference between tasks. Finding the optimal $\gamma$ requires careful tuning based on the specific tasks and model architecture. In \cref{sec:conclusion}, we discuss future directions for reducing feature interference beyond scaling.

\vspace{-0.2cm}

\section{Experiments}

\vspace{-0.2cm}

\subsection{Experimental Setup}\label{sec:exp_setup}

\paragraph{Evaluation Setup.} We evaluate {\method} across diverse fine-tuning methods, tasks, model architectures, and model size. For the fully fine-tuning, we evaluate NLP tasks on T5 following protocols of \citet{yadav2023ties}. For parameter-efficient fine-tuning (PEFT), we evaluate LoRA on GPT-2 for NLP tasks since it uses linear transformation matrices through low-rank adaptation, which aligns well with our approach, rather than vector parameters like IA3~\cite{liu2022few} in previous research.
To evaluate {\method} when model size scales up, we follow the setup of \citet{du2024parameter} to merge LLMs. Additionally, to evaluate the out-of-distribution robustness, we follow \citet{yadav2023ties} to test it on NLP tasks.

All experiments are conducted on NVIDIA A100 GPUs and AMD EPYC 7763 CPUs. {\method} is implemented in PyTorch and performed on NVIDIA A100 GPUs. The hyperparameter settings for {\method} are in \cref{app:hyperparameters}.

\paragraph{Baseline Methods.}
We compare {\method} against four established model merging approaches: (1) \textbf{Averaging} \cite{choshen2022fusing,wortsman2022model}, which computes the element-wise mean of individual models; (2) \textbf{Task Arithmetic} \cite{ilharco2022editing}, which merges by scaling and adding task vectors to the initial model;
(3) \textbf{Fisher Merging}~\cite{matena2022merging}, which approximates Fisher Information Matrix to weight parameters based on their importance for each task and combines the weighted parameters into the final merged model;
(4) \textbf{RegMean}~\cite{jin2022dataless}, which computes a closed-form solution to a least-squares regression problem that aims to minimize the distance between the merged model's activations and the individual models' activations.
(5) \textbf{Ties-Merging}, which enhances merging by eliminating redundant parameters and resolving sign conflicts; (6) \textbf{PCB-Merging}, which balances parameter competition through intra-task significance and inter-task similarity analysis. We also report results from individual \textbf{fine-tuned models} and a \textbf{pre-trained model} on all tasks; (7) \textbf{MetaGPT}~\cite{zhou2024metagpt}, which formalizes the objective of model merging into a multi-task learning framework, aiming to minimize the average loss difference between the merged model and each individual task model; (8) \textbf{Knots}~\cite{stoica2024model}, which utilizes Singular Value Decomposition to transform and align task-updates from multiple LoRA models into a shared representational space; (9) \textbf{Consensus TA}~\cite{wang2024localizing} and (10) \textbf{Localize-and-Stitch}~\cite{he2024localize}, both of which focus on constructing a sparse mask for each task to reduce inter-task conflict during merging.

\subsection{Results}

\begin{table*}[h]
  \centering
  \belowrulesep=0pt
  \aboverulesep=0pt
  \setlength{\tabcolsep}{2.5pt}
  \small
  \begin{tabular}{c|c|ccccccc}
      \toprule
      \rowcolor{mygray}  &       & \multicolumn{7}{c}{\textbf{Test Set Performance}} \\
      \rowcolor{mygray} \multirow{-2}{*}{\textbf{Method}} & \multirow{-2}{*}{\textbf{Average}} & paws & qasc & quartz & story\_cloze & wiki\_qa & winogrande & wsc \\
      \midrule
      Pre-trained & 53.5 & 49.9 & 35.8 & 53.3 & 48.1 & 76.2 & 50.0 & 61.1 \\
      Fine-tune & 81.0 & 91.4 & 95.5 & 79.1 & 79.6 & 95.2 & 62.8 & 63.6 \\
      \midrule
      Parameter Average & 60.2 & 55.2 & 57.5 & 55.2 & 49.4 & 91.1 & 50.2 & 62.5 \\
      Fisher Merging & 68.2 & 67.9 & 84.4 & 63.5 & 57.1 & 90.1 & 54.2 & 60.8 \\
      RegMean & 71.5 & 76.2 & 92.8 & 62.6 & 63.6 & 89.4 & 57.4 & 58.3 \\
      Task Arithmetic & 71.6 & 71.1 & 81.4 & 62.1 & 77.1 & 95.0 & 57.4 & 56.9 \\
      Ties-Merging & 72.2 & 78.8  & 88.4  & 65.1  & 70.7  & 84.2  & 56.2  & 62.3\\
      PCB-Merging & \underline{72.3} & 71.5 & 91.7 & 66.8 & 62.7 & 92.8 & 57.1 & 63.3 \\
      MetaGPT & 58.5 & 56.7 & 67.3 & 58.1 & 45.4 & 80.2 & 55.5 & 46.2 \\
      Consensus TA & 72.3 & 73.8 &	83.4 &	61.9 &	76.0 &	83.8 &	56.2 &	61.0 \\
      Localize-and-Stitch &	72.9 & 75.7 &	88.6 & 62.3 & 73.6 & 93.9 & 55.9 & 60.6 \\
      \textbf{{\method} (ours)} & \textbf{73.7} & 77.4 & 89.1 & 62.6 & 75.2 & 94.2 & 56.4 & 61.1 \\
      \bottomrule
  \end{tabular}
    \caption{Merge fully-fine-tuned T5-base model with different methods.}
  \label{tab:t5}
\end{table*}

\paragraph{Fully Fine-tuned NLP Model: T5 Merging}

For NLP experiments, we evaluate on T5-base \cite{colin2020exploring}, an encoder-decoder transformer \cite{vaswani2017attention} pre-trained with masked language modeling. We finetune T5-base on seven diverse tasks spanning question answering (QASC \cite{khot2020qasc}, WikiQA \cite{yang-etal-2015-wikiqa}, QuaRTz \cite{tafjord2019quartz}), paraphrase identification (PAWS \cite{paws2019naacl}), sentence completion (Story Cloze \cite{sharma2018tackling}), and coreference resolution (Winogrande \cite{sakaguchi2020winogrande}, WSC \cite{wsc}). We use the code from \citet{yadav2023ties} to finetune the model on these tasks.
To reduce variance and ensure reliable evaluation, we report the experimental results of the average performance over different templates~\cite{bach2022promptsource} for each task.

As shown in \cref{tab:t5}, {\method} achieves state-of-the-art performance when merging T5-base models, outperforming existing methods by $1.4\%$ on average across 7 tasks. {\method} shows particularly strong performance on PAWS and Story Cloze, with improvements of $5.9\%$ and $12.5\%$ respectively over PCB-Merging, the previous best method.

\paragraph{PEFT of NLP Model: LORA Adapters Merging}
For PEFT experiments, we evaluate GPT-2 Medium model on LoRA adapters \cite{hu2021lora}, which are task-specific adapters that are fine-tuned on NLP task datasets. These tasks include converting tables (E2E\cite{novikova2017e2e}), knowledge graph (WebNLG\cite{gardent2017webnlg}) and structured data (DART\cite{nan2020dart}) to natural language. We use the released checkpoints from \citet{hu2021lora} and evaluate the performance of the merged model on these datasets. Specifically, we compare Knots, which is designed specifically for LoRA merging.

As shown in \cref{tab:lora}, {\method} outperforms existing baselines on all metrics. Specifically, comparing with previous best baselines, {\method} shows improvements of $2.2\%$ for NIST of E2E, $4.8\%$ over PCB-Merging for CIDEr of E2E.

\begin{table*}[h!]
  \centering
  \belowrulesep=0pt
  \aboverulesep=0pt
  \small
  \setlength\tabcolsep{2pt}
      \begin{tabular}{c|ccccc|ccc|ccc|c}
          \toprule
          \rowcolor{mygray}        &  \multicolumn{11}{c|}{\textbf{Test Set Performance}} & \\
          \rowcolor{mygray}        &  \multicolumn{5}{c|}{\textbf{E2E}} &  \multicolumn{3}{c|}{\textbf{DART}} &  \multicolumn{3}{c|}{\textbf{WebNLG}} & \\ 
          \rowcolor{mygray} \multirow{-2}{*}{\textbf{Method}  } & {BLEU$\uparrow$} & {NIST$\uparrow$} & {MET$\uparrow$} & {ROUGE-L$\uparrow$} & {CIDEr$\uparrow$} & {BLEU$\uparrow$}  & {MET$\uparrow$} & {TER$\downarrow$} & {BLEU-A$\uparrow$} & {MET-A$\uparrow$} & {TER-A$\downarrow$} &  \multirow{-2}{*}{\textbf{Rank}}\\
          \midrule
          Pre-trained & 0.2 & 0.58 & 1.5 & 5.2 & 0.002 & 0.2 & 2.0 & 152.4 & 0.15 & 2.0 & 179.1 & -- \\
          LoRA & 67.7 & 8.64 & 46.0 & 68.3 & 2.36 & 44.8 & 35.0 & 50.4 & 52.3 & 37.0 & 44.4 & -- \\
          \midrule
          Parameter Average & 63.4 & 8.00 & 40.8 & 66.6 & 2.01 & 40.0 & 32.0 & 53.7 & 43.9 & 32.0 & 49.2 & 7 \\
          Task Arithmetic & 63.2 & 8.02 & 40.9 & 66.3 & 1.98 & 40.8 & 33.0 & 53.7 & 45.9 & 34.0 & 48.8 & 6 \\
          Ties-Merging & 62.8 & 8.14 & 41.1 & 65.9 & 2.08 & 41.4 & 33.0 & 53.7 & 46.1 & 33.0 & 48.1 & 5 \\
          PCB-Merging & 62.9 & 8.12 &41.4 & 66.0 & 2.08 & 41.3 & 33.0 & 53.5 & 46.2 & 33.0 & 48.1 & 3 \\
          Knots & 62.9 & 8.22 & 41.6 & 65.8 & 2.05 & 41.52 & 34 & 54.0 & 46.0 & 34 & 47.5 & 2 \\
          MetaGPT & 63.4 & 7.88 & 40.7 & 66.1 & 2.01 & 42.23 & 33 & 52.1 & 45.9 & 33 & 47.4 & 4 \\
          \textbf{{\method} (ours)} & 64.1 & 8.40 & 42.2 & 66.5 & 2.18 & 41.6 & 33.0 & 54.1 & 47.1 & 34.0 & 48.0 & 1 \\
      
          \bottomrule
      \end{tabular}
\caption{Merge LoRA Adapters of GPT-2 M with Different Methods. $\uparrow$ indicates higher is better, $\downarrow$ indicates lower is better.}
\label{tab:lora}
\end{table*}

\begin{table}[ht]
  \belowrulesep=0.8pt
  \aboverulesep=0.8pt
  \centering
  \setlength{\tabcolsep}{1.8pt}
  \small
  \begin{tabular}{c|c c c|c}
  \toprule
  \rowcolor{mygray}        &  \multicolumn{3}{c|}{\textbf{Datasets}} & \\
  \rowcolor{mygray} \multirow{-2}{*}{\textbf{Method}  }        &  CMMLU &  GSM8K &  Humaneval &  \multirow{-2}{*}{\textbf{Average}} \\
  \midrule
   Chinese          & 38.6  & 2.3   & 13.4      & 18.1    \\
  Math            & 31.2  & 65.6  & 0       & 32.3    \\
  Code            & 33.3  & 0   & 17.1      & 16.8    \\
  \midrule
  Parameter Average         & 35.6  & 48.5  & 6.7      & 30.3    \\
  Task Arithmetic & 35.4  & 46.1  & 9.8       & 30.4    \\
  Ties-Merging      & 36.5  & 53.4  & 12.8      & 34.3    \\
  MetaGPT &  36.2 & 50.6  &  16.9     & 34.6    \\
  PCB-Merging      & 36.4  & 52.3  & 16.5      & {35.1}    \\
  \textbf{{\method} (ours)}            & 36.5  & 63.0  & 14.0      & \textbf{37.8} \\
  \bottomrule
  \end{tabular}
  \caption{Results on the CMMLU, GSM8K, and Humaneval datasets.}\label{tab:llm}
\end{table}

\paragraph{Large Model: LLM Merging}
We evaluate model merging on three fine-tuned Llama-2-7B models~\cite{touvron2023llama} focusing on different capabilities: Chinese language proficiency\footnote{\url{https://huggingface.co/LinkSoul/Chinese-Llama-2-7b}}, mathematical reasoning~\cite{yu2023metamath}\footnote{\url{https://huggingface.co/meta-math/MetaMath-7B-V1.0}}, and code generation~\cite{roziere2024code}\footnote{\url{https://huggingface.co/qualis2006/llama-2-7b-int4-python-code-18k}}. Each model's performance was assessed using domain-specific benchmarks: CMMLU~\cite{li2024cmmlu} for Chinese language understanding, GSM8K~\cite{cobbe2021training} for mathematical reasoning, and HumanEval~\cite{chen2021evaluating} for code generation abilities.

As shown in \cref{tab:llm}, {\method} achieves state-of-the-art performance across all three domains, improving overall performance by $2.7\%$ compared to the best baseline. The most significant improvement is observed in mathematical reasoning, where {\method} outperforms other methods by approximately $10\%$ on the GSM8K benchmark.

\paragraph{Out-of-Distribution Generalization}
To evaluate out-of-distribution generalization, we test the merged T5-base model (previously trained on seven tasks) on six held-out tasks from the T0 mixture~\cite{sanh2021multitask}: three question answering tasks (Cosmos QA \cite{huang2019cosmos}, Social IQA \cite{sap2019social}, QuAIL \cite{quail_dataset}), word sense disambiguation (WiC \cite{pilehvar2018wic}), and two sentence completion tasks (COPA \cite{copa}, H-SWAG \cite{zellers2019hellaswag}). As shown in \cref{tab:ood}, {\method} achieves $0.9\%$ improvement over the best baseline on T5-base, demonstrating strong generalization capabilities to tasks outside the training distribution.

\subsection{Additional Results and Analysis}\label{ssec:additional_results}

\begin{figure*}[th]
    \centering
    \vspace{-10pt}
    \subfigure[Merge task matrices vs fine-tuned matrices]{
    \includegraphics[width=0.23\textwidth]{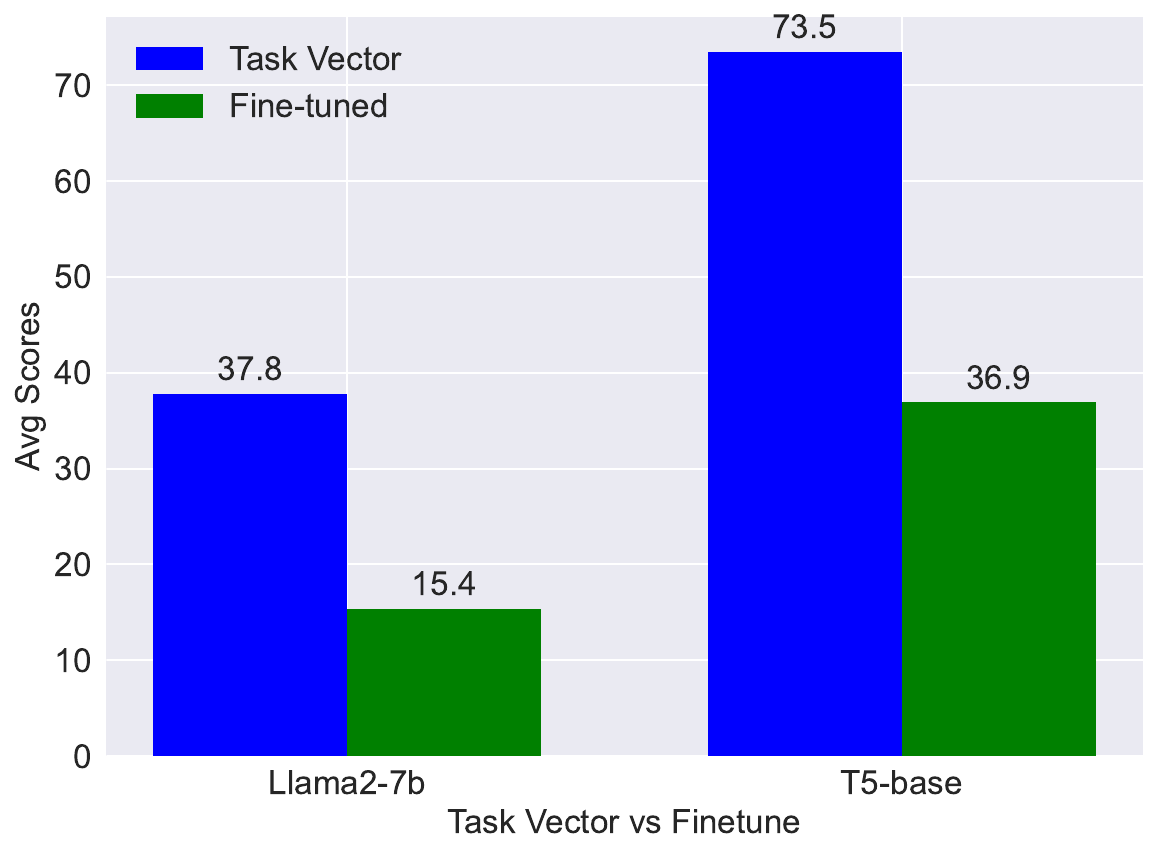}
    \label{fig:task-vector-vs-finetune}
 }
 \hspace{-3pt}
    \subfigure[Performance vs number of tasks and scaling factor]{
    \includegraphics[width=0.23\textwidth]{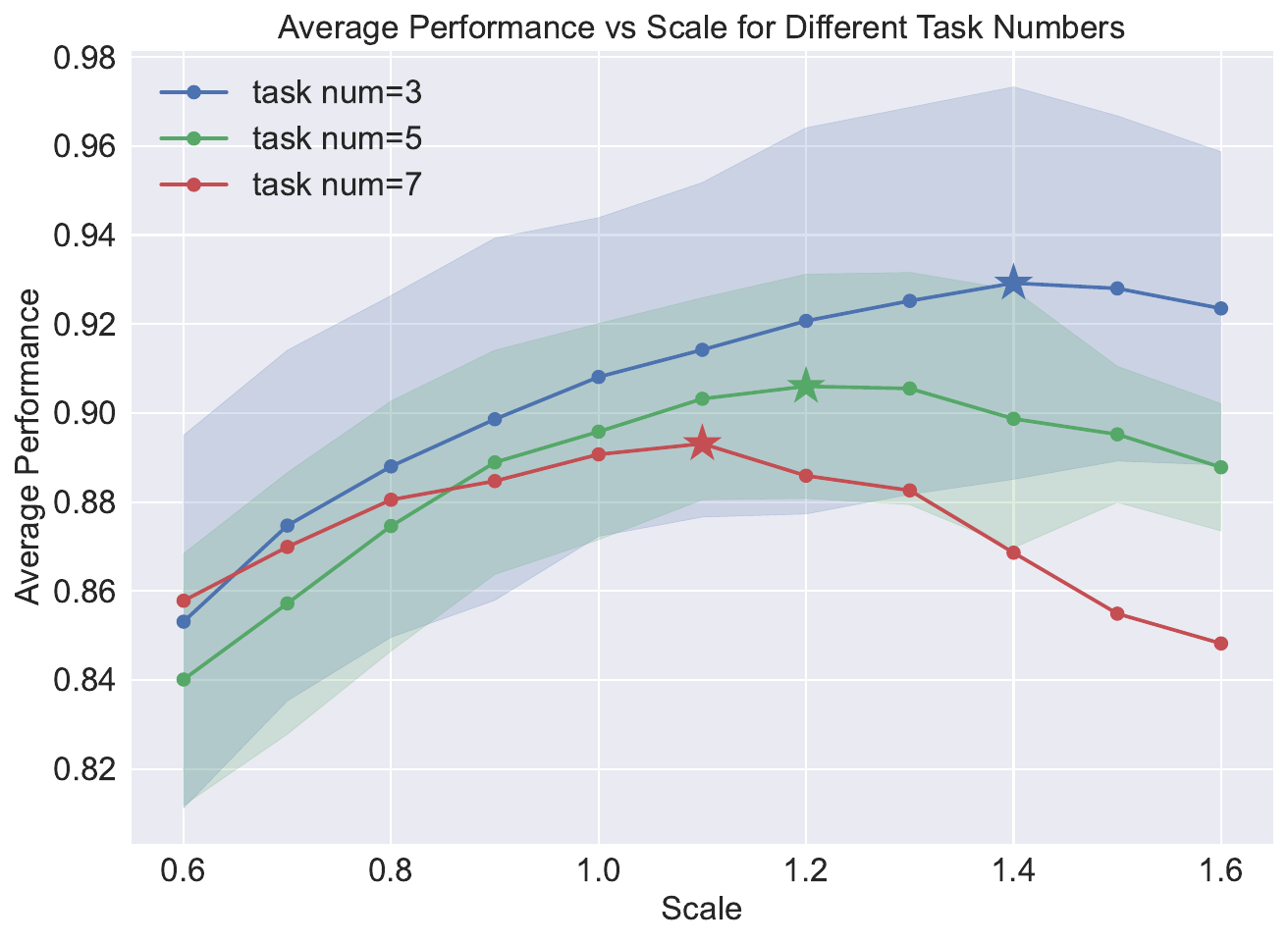}
    \label{fig:feature-interference}
 }
 \hspace{-10pt}
    \subfigure[Various $\eta$ at $\gamma$ = 0.6]{
    \includegraphics[width=0.23\textwidth]{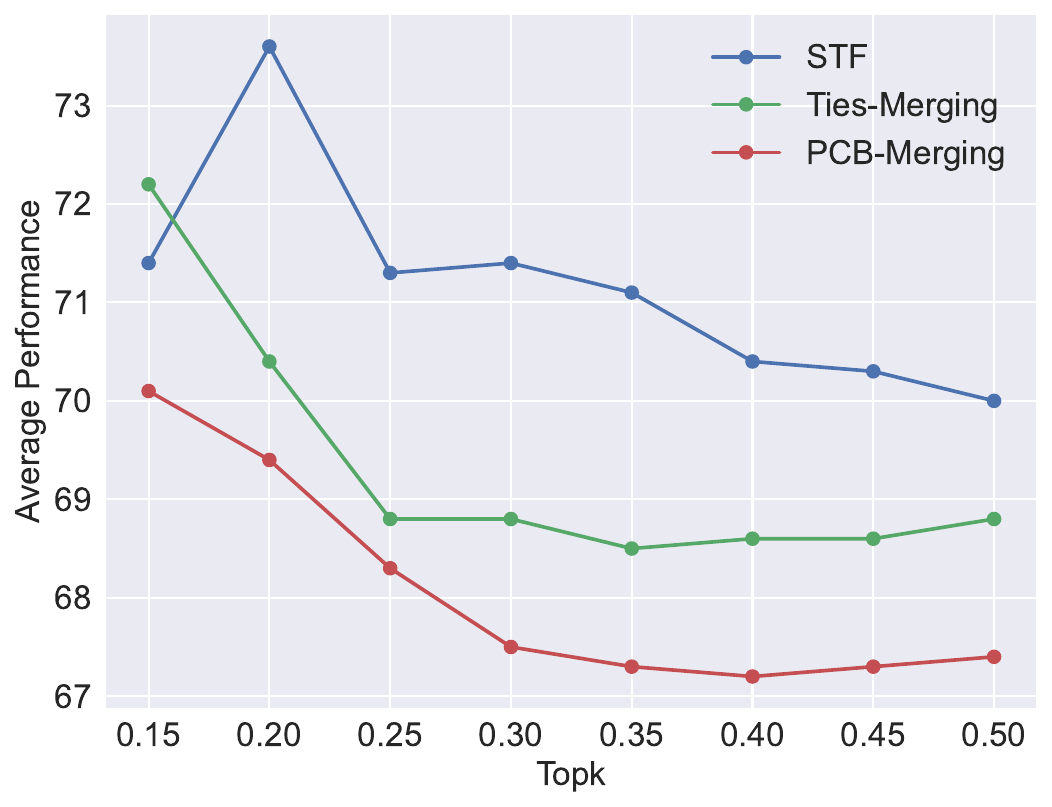}
    \label{fig:hyperparameter-sensitivity-eta}
 }
 \hspace{-12pt}
    \subfigure[Various $\gamma$ at $\eta$ = 0.2]{
    \includegraphics[width=0.23\textwidth]{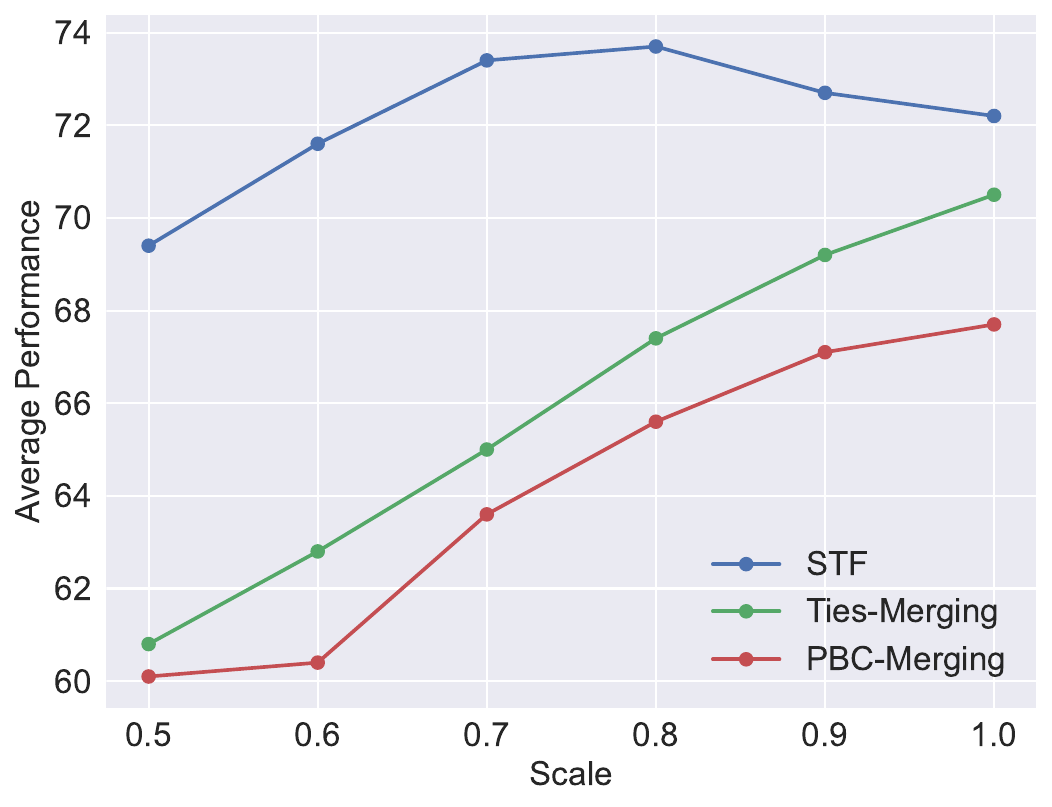}
    \label{fig:hyperparameter-sensitivity-gamma}
    }
    \vspace{-10pt}
    \caption{Experimental results on analyzing {\method}.}
    \vspace{-10pt}
\end{figure*}

\paragraph{Task Matrices versus Fine-tuned Matrices} We compare merging task matrices versus fine-tuned matrices directly on T5-base models. As shown in \cref{fig:task-vector-vs-finetune}, merging task matrices consistently outperforms merging fine-tuned matrices across all architectures by a large margin. This validates our discussion in \cref{sec:discussion} that task matrices better isolate task-specific features compared to fine-tuned matrices, making them more effective for model merging.

\vspace{-0.2cm}

\paragraph{Feature Interference and Scaling Factor} We examine how performance of T5 merging varies with different number of tasks and scaling factor in \cref{fig:feature-interference}. We observe that as more tasks are merged, the optimal scaling factor $\gamma$ that achieves the best performance decreases. This trend indicates that larger $\gamma$ values amplify interference between task-specific features when merging many tasks, requiring smaller scaling factors to maintain performance. The results validate our discussion in \cref{sec:discussion} about scaling merged matrix to balance feature preservation against interference.

\paragraph{Hyperparameter Sensitivity}
We analyze the sensitivity of {\method} to the hyperparameters $\eta$ and $\gamma$ on the T5-base. 
Three model merging methods are compared in the experiments: {\method}, Ties-Merging, and PCB-Merging. We follow the experiment setup of previous work \cite{du2024parameter} to analyze the effect of two parameters.
The first parameter is the scaling factor $\gamma$ , which is used to scale the merged task vector. For {\method} and Ties-Merging, the second hyperparameter is the top-k percentage $\eta$, indicating that the $\eta\%$ parameters in task vector with the highest absolute value are retained. For PCB-Merging, the second hyperparameter is mask ratio $\eta$, indicating that the $\eta\%$ parameters in parameter competition balancing matrix with the biggest absolute value will be retained.  
We vary the hyperparameters $\gamma$ with step size 0.1 from 0.5 to 1.0, and $\eta$ with step size 0.05 from 0.15 to 0.5.

First, in \cref{fig:hyperparameter-sensitivity-eta,fig:hyperparameter-sensitivity-gamma}, {\method} achieves optimal performance with $\gamma = 0.8$ and $\eta = 0.2$ on T5-base.
Second, performance varies with both hyperparameters. For sparsity ratio $\eta$, performance first improves then declines as it increases. When $\eta$ is too low, important features are filtered out as most parameters are set to zero. When $\eta$ is too high, noise in the task matrices affects the quality of singular vectors. For scaling factor $\gamma$, we observe a similar trend - performance initially improves then deteriorates with increasing values. A small $\gamma$ diminishes the magnitude of singular vectors, while a large $\gamma$ may amplify interference between tasks.

\paragraph{Model Merging Time} 
We analyze the time required for model merging with {\method} on different models. As shown in \cref{tab:time-comparison}, the merging time increases with model size, primarily due to the SVD computation on larger matrices.
For smaller models, TIES and PCB are more efficient than Knots and {\method}. However, with larger models, {\method} becomes more efficient because TIES and PCB require storing all parameters in memory simultaneously and run only on CPU, while {\method} processes one task matrix at a time and leverages GPU acceleration.
The relatively short merging times demonstrate that {\method} is practical for real-world applications, despite its theoretical time complexity.

\begin{table}[h]
  \belowrulesep=0.8pt
  \aboverulesep=0.8pt
    \centering
  \setlength{\tabcolsep}{2.5pt}
  \small
    \begin{tabular}{lccccc}
        \toprule
        \rowcolor{mygray} Method & T5-base & GPT-2 LoRA & Llama-2-7B \\
        \midrule
        TIES & 34 & 0.08 & 1280 \\
        PCB & 113 & 0.09 & 1593 \\
        Knots & -- & 8 & -- \\
        {\method} (Ours) & 127 & 5 &  623\\
        \bottomrule
    \end{tabular}
    \caption{Merging time for different models (unit: s). We run PCB and TIES on CPU because of the large memory requirement.}\label{tab:time-comparison}
\end{table}

\vspace{-10pt}

\section{Conclusion}\label{sec:conclusion}
In this paper, we present {\method}, a novel model merging approach that preserves task-specific features in linear transformations through feature superposition. Extensive experiments demonstrate that {\method} consistently outperforms existing methods across different architectures and tasks.
The success of {\method} demonstrates the value of leveraging working mechanisms of deep neural networks in model merging, rather than treating them at the parameter level.


\section*{Limitations}

While {\method} effectively preserves features during merging, it does not explicitly identify task-specific features corresponding to semantic meanings. Future work could leverage mechanistic interpretability techniques like superposition analysis and sparse autoencoders to better isolate and preserve task-specific features in linear transformations. Additionally, {\method} currently relies on a simple scaling approach to manage interference between tasks. More sophisticated methods for analyzing feature importance and selectively removing interference from less critical features could further improve performance. These advances would help develop a more principled approach to preserving task capabilities during model merging. Another major limitation is the time complexity of SVD and solving the linear system in {\method}, which can be improved by using more efficient method for feature extraction in the future.

\section*{Acknowledgements}

This work is supported by 
National Key Research and Development Program of China (Grant No.2023YFB2903904), 
National Natural Science Foundation of China (Grant No. 92270106), 
Beijing Natural Science Foundation (Grant No. 4242039),
and CCF-Huawei Populus Grove Fund.

\bibliography{custom}

\appendix

\onecolumn

\section{Measure Feature Preservation}\label{app:feature_preservation}

We measure feature preservation by $\left| \langle \sigma_i^{(k)} \bfu_{i}^{(k)}, \bar{\MM} \bfv_{i}^{(k)} - \sigma_i^{(k)} \bfu_i^{(k)}\rangle \right|$, where $\sigma_i^{(k)} \bfu_{i}^{(k)}$ and $\bar{\MM} \bfv_{i}^{(k)}$ and $\bfv_{i}^{(k)}$ are the task-specific features identified in \cref{ssec:feature-superposition}, and $\bar{\MM}$ is the merged task matrix for various merging methods without scaling. This measurement is based on the superposition objective in \eqref{eq:obj}. Higher values indicate that the output of merged task matrix $\bar{\MM}$ is less similar to the output feature vectors of original task matrix $\TM$, and thus the feature preservation is worse.

\begin{figure}[h]
    \centering
    \includegraphics[width=0.80\textwidth]{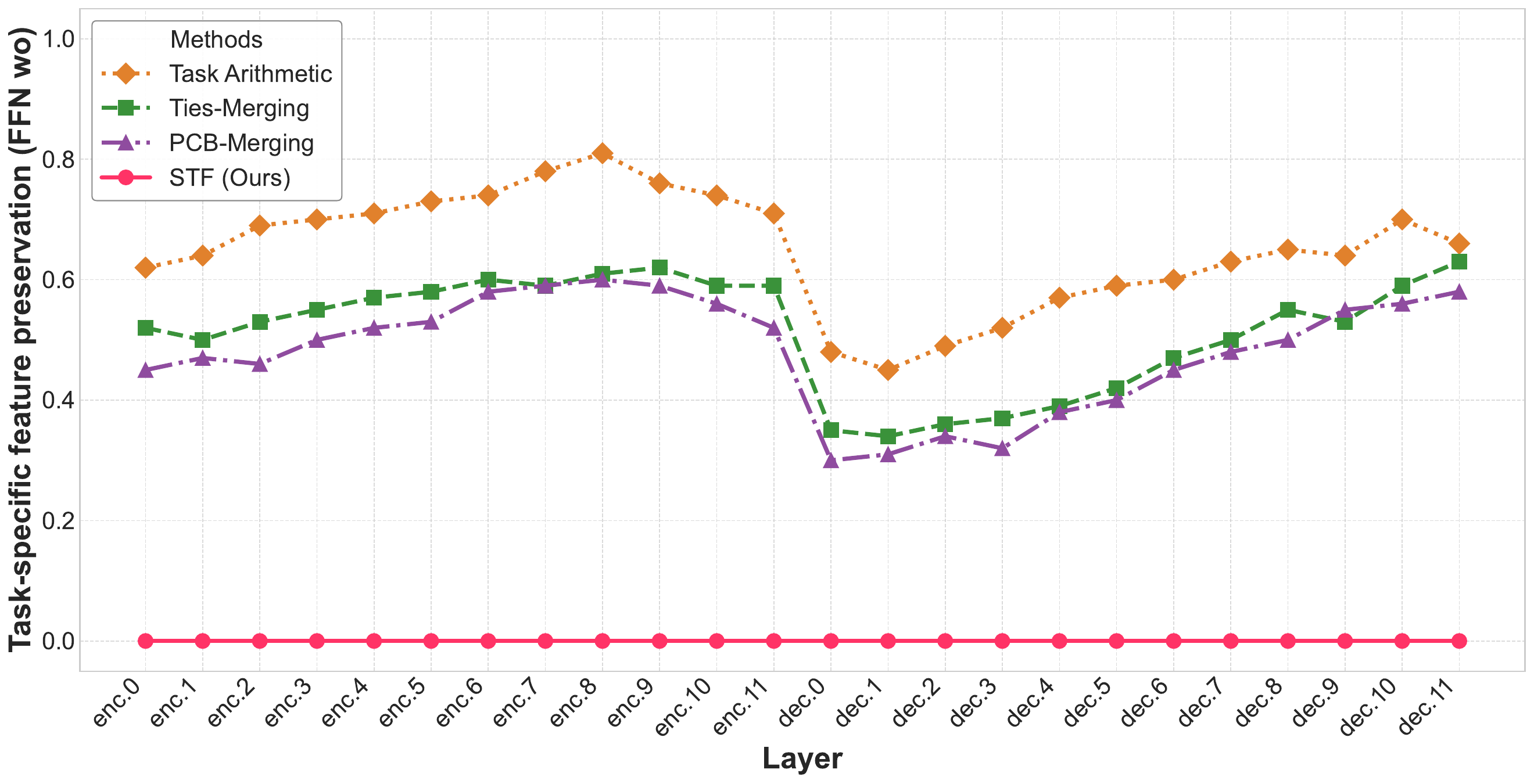}
    \caption{Feature preservation for FFN output layer in T5-base model}
    \label{fig:feature-preservation}
    \vspace{-10pt}
\end{figure}

We calculate the preservation for every feature vector in the original fine-tuned models across all datasets, and then average all these individual preservation values. In \cref{fig:abs_direction_mismatch_ffn_wi_overall_avg}, we choose the input linear transformation matrices of all FFN layers in T5-base model. We also measured the feature preservation for other layers, including the attention layers. \cref{fig:feature-preservation} shows the feature preservation for the output layers of FFN in T5-base model, which shows the same trend with \cref{fig:abs_direction_mismatch_ffn_wi_overall_avg}. We observe that {\method} exactly preserves the magnitudes and directions of output feature vectors, while Ties-Merging and PCB-Merging have much lower feature preservation. The results indicate that better feature preservation strongly correlates with better performance and validate our hypothesis that feature superposition is effective for model merging.

\section{Singular Value Decomposition of Task Matrices}\label{app:svd}

Applying SVD to task matrix, we have $\TM=\sum_{k=1}^{r_i} \sigma_{i}^{(k)} \bfu_{i}^{(k)} \bfv_{i}^{(k)\top}$, where $\sigma_{i}^{(k)}$ is the $k$-th singular value, and $\bfu_{i}^{(k)}$ and $\bfv_{i}^{(k)}$ are the $k$-th left and right singular vectors of $\TM$, respectively. These singular vectors serve as orthogonal basis for output $\TM \mathbf{x}$ and input $\mathbf{x}$. Specifically, any input representation $\mathbf{x}\in \Real^n$ can be decomposed as $\mathbf{x} = \sum_{k=1}^{r_{i}} w_{i,j}^{(k)} \bfv_{i}^{(k)} + \bfb$, where $\bfb$ is orthogonal to all right singular vectors. When $\TM$ transforms $\mathbf{x}$, we get $\TM \mathbf{x} = \sum_{k=1}^{r_{i}} \sigma_{i}^{(k)} w_{i,j}^{(k)} \bfu_{i}^{(k)}$ since $\bfb$ lies in the null space of $\TM$. This shows that the input $\mathbf{x}$ and output $\TM \mathbf{x}$ can be represented as a weighted sum of right and left singular vectors, respectively.

\section{Proof of \cref{thm:linear}}\label{app:linear}
Here we use $\mathbf{X} = [\mathbf{x}_{i}^{(k)\top} \mathbf{x}_{i'}^{(k')}]_{i,k;i',k'}$ to denote the matrix formed by enumerating over $i,k$ for the row index and $i',k'$ for the column index. 

\begin{proof}
The objective of feature superposition can be converted to
\begin{equation*}
    \left<\bfu_{i}^{(k)}, \MM \bfv_{i}^{(k)} \right> = \left<\bfu_{i}^{(k)}, \TM \bfv_{i}^{(k)}\right>, \quad \forall i,k.
\end{equation*}

We bring $\MM=
\sum\nolimits_{i'=1}^T 
\sum\nolimits_{k'=1}^{r_{i}} \alpha_{i'}^{(k')} \bfu_{i'}^{(k')} \bfv_{i'}^{(k') \top}$ and $\TM=\sum\nolimits_{k'=1}^{r_{i}} \alpha_{i}^{(k')} \bfu_{i}^{(k')} \bfv_{i}^{(k') \top}$ to the left side of the equation, and move $\sigma_{i}^{(k)}$ to the right side.

Then, the objective becomes
\begin{align}\label{eq:linear}
\bfu_{i}^{(k)\top} \!\!\left( \sum_{i'=1}^T \sum_{k'=1}^{r_{i}} \!\alpha_{i'}^{(k')} \bfu_{i'}^{(k')} \bfv_{i'}^{(k') \top}\!\! \right)\! \bfv_{i}^{(k)} \!=\! \sigma_{i}^{(k)} 
     \Leftrightarrow  \sum_{i'=1}^T \sum_{k'=1}^{r_{i}}  \!\! \bfu_{i}^{(k)\top} \bfu_{i'}^{(k')} \bfv_{i'}^{(k') \top} \bfv_{i}^{(k)} \alpha_{i'}^{(k')} \!=\! \sigma_{i}^{(k)}
\end{align}
which is one linear equation with $\bfu_{i}^{(k)\top} \bfu_{i'}^{(k')} \bfv_{i'}^{(k') \top} \bfv_{i}^{(k)}$ as the coefficient of variable $\alpha_{i'}^{(k')}$. We can convert all equations in \eqref{eq:obj} to a linear system for all tasks in $\{\mathsf{T}_1,\ldots, \mathsf{T}_T\}$:
\begin{equation*}
    \begin{bmatrix}
        \bfu_{1}^{(1)\top} \bfu_{1}^{(1)} \bfv_{1}^{(1) \top} \bfv_{1}^{(1)} & \cdots & \bfu_{1}^{(1)\top} \bfu_{T}^{(r_{T})} \bfv_{T}^{(r_{T}) \top} \bfv_{1}^{(1)} \\
        \vdots & \ddots & \vdots \\
        \bfu_{T}^{(r_{T})\top} \bfu_{1}^{(1)} \bfv_{1}^{(1) \top} \bfv_{T}^{(r_{T})} & \cdots & \bfu_{T}^{(r_{T})\top} \bfu_{T}^{(r_{T})} \bfv_{T}^{(r_{T}) \top} \bfv_{T}^{(r_{T})}
    \end{bmatrix}
    \begin{bmatrix}
        \alpha_{1}^{(1)} \\
        \vdots \\
        \alpha_{T}^{(r_{T})}
    \end{bmatrix}
    =
    \begin{bmatrix}
        \sigma_{1}^{(1)} \\
        \vdots \\
        \sigma_{T}^{(r_{T})}
    \end{bmatrix}.
\end{equation*}
With Kronecker product, we can write the linear system in a more compact form:
\begin{equation*}
    \SigM \circ \UM \circ \VM \boldsymbol{\alpha} = \boldsymbol{\sigma},
\end{equation*}
where $\UM = [\bfu_{i}^{(k)\top} \bfu_{i'}^{(k')}]_{i,k;i',k'}=\bar{\mathbf{U}}^\top \bar{\mathbf{U}}\in \Real^{r\times r}$;  $\VM = [\bfv_{i'}^{(k') \top} \bfv_{i}^{(k)}]_{i',k';i,k}=\bar{\mathbf{V}}^\top \bar{\mathbf{V}} \in \Real^{r\times r}$; $\boldsymbol{\alpha}$ is the vector of variables $\alpha_{i'}^{(k')}$; and $\boldsymbol{\sigma}$ is the vector of singular values $\sigma_{i}^{(k)}$. The index $i,k,i',k'$ share consistent order across $\UM$ and $\VM$. 
\end{proof}

\section{More Experimental Results}\label{app:more_results}

\begin{table*}[h]
    \centering
    \belowrulesep=0pt
    \aboverulesep=0pt
    \caption{Out-of-Distribution Generalization Performance of merged T5-base model}
    \label{tab:ood}
    \begin{tabular}{c|c|cccccc}
        \toprule
        \rowcolor{mygray}  &       & \multicolumn{6}{c}{\textbf{Out-of-Distribution Performance}} \\
        \rowcolor{mygray} \multirow{-2}{*}{\textbf{Method}} & \multirow{-2}{*}{\textbf{Average}} & cosmos\_qa & social\_iqa & quail & wic & copa & h-swag \\
        \midrule
        Pre-trained & 31.1 & 21.9 & 18.8 & 24.1 & 65.6 & 43.8 & 12.5 \\
        \midrule
        Average & 36.3 & 23.5 & 37.0 & 25.4 & 50.2 & 54.5 & 27.2 \\
        Task Arithmetic & 37.0 & 21.9 & 36.8 & 25.5 & 49.5 & 61.4 & 26.6 \\
        Ties-Merging & 37.1 & 22.2 & 37.8 & 24.9 & 51.6 & 62.2 & 26.5 \\
        PCB-Merging & \underline{37.3} & 23.0 & 38.1 & 24.6 & 52.2 & 59.0 & 27.2 \\
        \textbf{{\method} (ours)} & \textbf{38.2} & 22.4 & 38.0 & 26.0 & 51.6 & 64.1 & 27.0  \\
        \bottomrule
    \end{tabular}
\end{table*}

\subsection{More Analysis}

\begin{figure*}[h!]
    \centering
    \vspace{-10pt}

    \subfigure[Remove smallest singular vectors]{
    \includegraphics[width=0.30\textwidth]{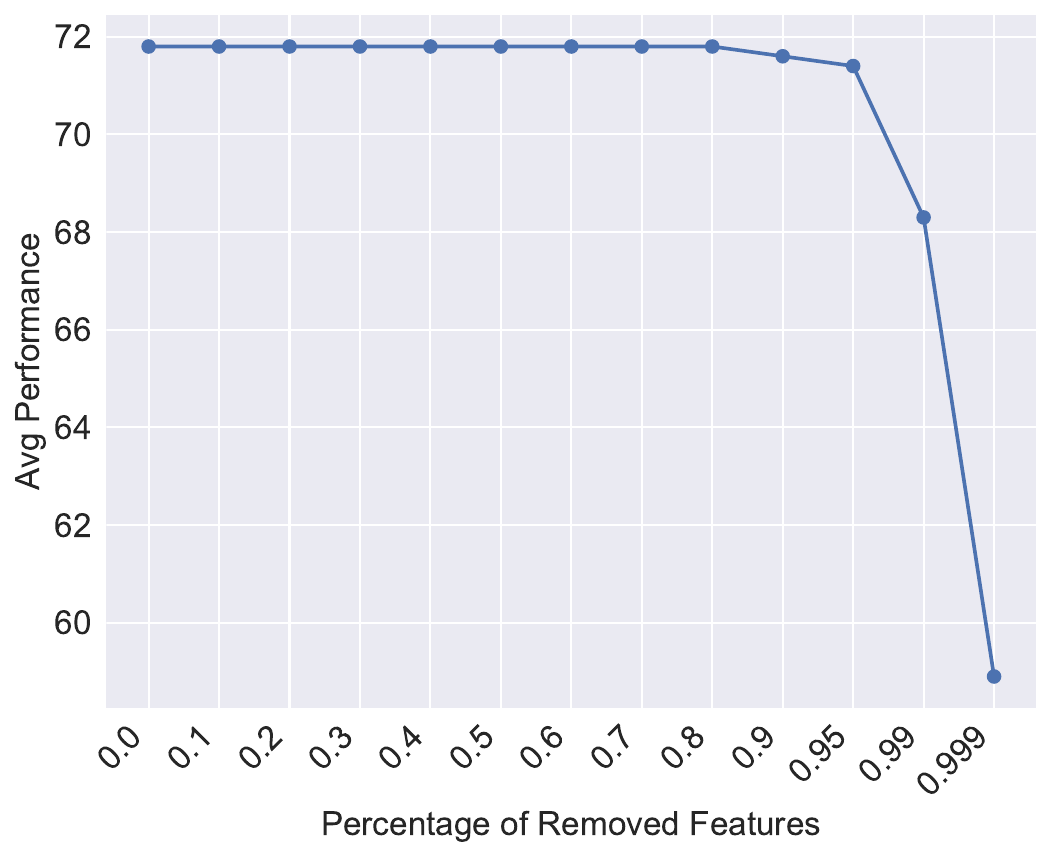}
    \label{fig:remove-small-feature}
 }
    \hspace{-10pt}
    \subfigure[Remove biggest singular vectors]{
    \includegraphics[width=0.30\textwidth]{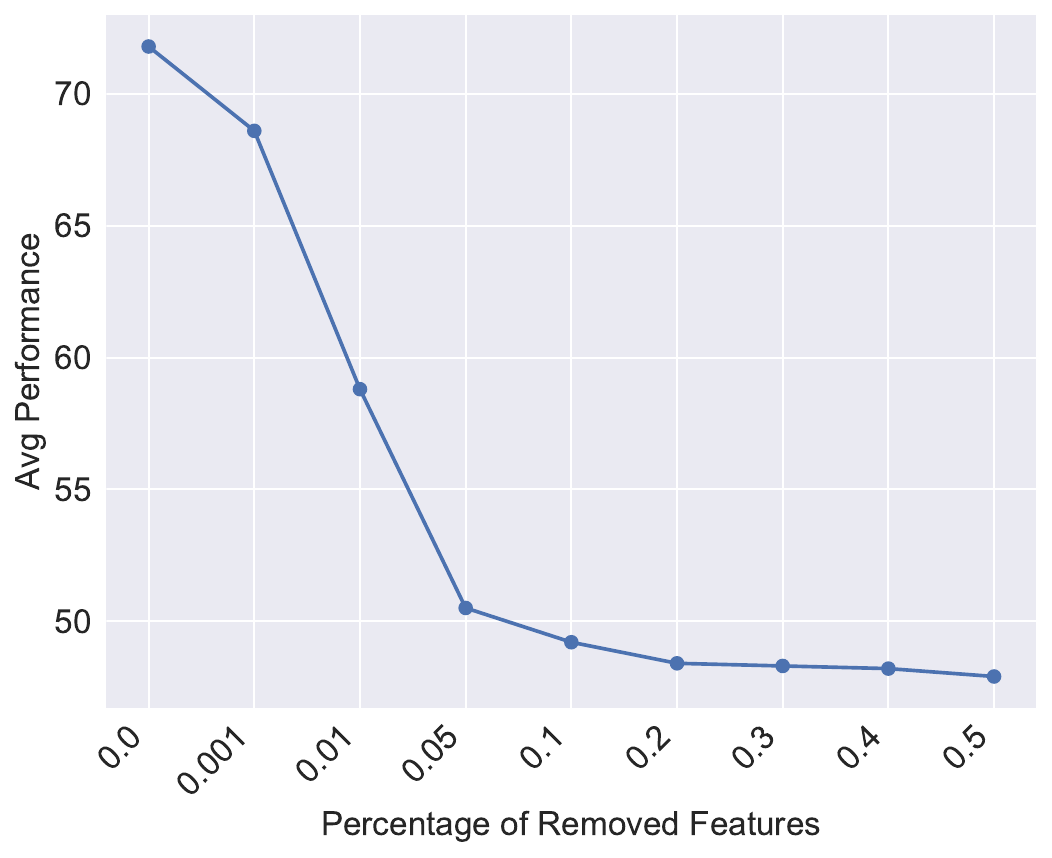}
    \label{fig:remove-big-feature}
 }
    \vspace{-10pt}
    \caption{Performance of {\method} with different percentage of removed features.}
    \vspace{-10pt}

\end{figure*}

\paragraph{Removing Singular Vectors}
We investigate removing singular vectors from task matrices before merging, analyzing on T5-base: removing smallest versus largest singular vectors. As shown in \cref{fig:remove-big-feature,fig:remove-small-feature}, removing up to 80\% of smallest singular vectors has minimal impact on performance, while removing just 5\% of largest singular vectors causes significant degradation. This indicates that while most singular features are redundant, preserving those with largest singular values is crucial for maintaining model capabilities. This insight could guide future work on selective feature preservation during model merging.

\paragraph{Fully Fine-tuned Vision Model: ViT Merging}
We evaluate {\method} on vision tasks using two CLIP models \cite{radford2021learning} with ViT-B/32 and ViT-L/14 architectures \cite{dosovitskiy2020image} as visual encoders. We use the released checkpoints from \citet{ilharco2022editing} that were fine-tuned on eight diverse classification tasks spanning remote sensing, traffic signs, and satellite imagery domains. These tasks include: Cars \cite{cars}, DTD \cite{dtd}, EuroSAT \cite{eurosat}, GTSRB \cite{gtsrb}, MNIST \cite{lecun1998mnist}, RESISC45 \cite{cheng2017remote}, SUN397 \cite{sun397}, and SVHN \cite{svhn}. 
For all experiments, we keep the text encoder fixed and only merge the parameters of visual encoders.

As shown in 
\cref{tab:vit}, 
{\method} achieves state-of-the-art performance when merging fully fine-tuned ViT models, outperforming existing methods by 1.1\% and 0.3\% on average across 8 tasks for ViT-B/32 and ViT-L/14 architectures respectively. For ViT-B/32, {\method} shows particularly strong performance on RESISC45 and EuroSAT, with 2.9\% and 11.1\% improvements over the best baseline. For ViT-L/14, {\method} maintains high performance across all tasks, demonstrating effective preservation of task-specific features during merging.

\begin{table*}[th]
  \centering
  \belowrulesep=0pt
  \aboverulesep=0pt
  \setlength{\tabcolsep}{4.5pt}
  \small
  \caption{\label{tab:app_vit_base} Test set performance when merging ViT-B/32 and ViT-L/14 models on 8 vision tasks.}\label{tab:vit}
  \begin{tabular}{c|r|c|cccccccc}
  
  \toprule
  \rowcolor{mygray} & \textbf{Task($\rightarrow$)} &    & \multicolumn{8}{c}{\textbf{Test Set Performance}} \\ 
  \rowcolor{mygray}\multirow{-2}{*}{\textbf{Model}}  & \textbf{Method($\downarrow$)} &  \multirow{-2}{*}{\textbf{Average}}  & SUN397  &  Cars  &  RESISC45  &  EuroSAT  &  SVHN  &  GTSRB  &  MNIST & DTD \\
  \midrule

  
  \multirow{7}{*}{{ViT-B/32}} &  {Pre-trained}  & 46.3 & 61.7 & 54.7 & 58.5 & 51.2 & 29.1 & 27.4 & 45.6 & 42.1 \\
   & {Fine-tuned}  & 90.5  &  75.3  &  77.7  &  96.1  &  99.7  &  97.5  &  98.7  &  99.7  &  79.4 \\
  \cmidrule{2-11}
 &  {Averaging} & 65.8  &  65.3  &  63.4  &  71.4  &  71.7  &  64.2  &  52.8  &  87.5  &  50.1 \\
 & {Fisher Merging} & 68.3 & \textbf{68.6} & \textbf{69.2} & 70.7 & 66.4 & 72.9 & 51.1 & 87.9 & 59.9 \\
 & {RegMean} & 71.8 & 65.3 & 63.5 & 75.6 & 78.6 & 78.1 & 67.4 & 93.7 & 52.0 \\
  &  {Task Arithmetic} & 70.1 &  63.8  &  62.1  &  72.0  &  77.6  &  74.4  &  65.1  &  94.0  &  52.2 \\
    & {Ties-Merging} & 73.6 & 64.8 & 62.9 & 74.3 & 78.9 & 83.1 & 71.4 & 97.6 & 56.2 \\
    & {PCB-Merging} & \underline{76.3} & 66.7 & 65.5 & 78.5 & 79.3 & \textbf{86.4} & \textbf{77.1} & \textbf{98.2} & 59.1 \\
    & \textbf{{\method} (ours)} & \textbf{77.4}    & 65.4	& 62.0	& \textbf{81.4}	& \textbf{90.4}	& {84.1}	& \textbf{77.1}	& 95.6	& \textbf{62.9}  \\
  \midrule
  \multirow{7}{*}{{ViT-L/14}} &  {Pre-trained} & 64.1 & 68.2 & 76.4 & 69.7 & 64.7 & 60.4 & 49.4 & 67.9 & 56.3 \\ 
& {Fine-tuned} & 94.2  &  82.3  &  92.4  &  97.4  &  100  &  98.1  &  99.2  &  99.7  &  84.1  \\
\cmidrule{2-11}
& {Averaging} & 79.6  &  72.1  &  81.6  &  82.6  &  91.9  &  78.2  &  70.7  &  97.1  &  62.8  \\
& Fisher Merging & 82.2 &  69.2  &  \textbf{88.6}  &  87.5  &  93.5  &  80.6  &  74.8  &  93.3  &  70.0  \\
& {RegMean} & 83.7 & 73.3 & 81.8 & 86.1 & \textbf{97.0} & 88.0 & 84.2 & 98.5 & 60.8 \\
& {Task Arithmetic} & 84.5 &  74.1  &  82.1  &  86.7  &  93.8  &  87.9  &  86.8  &  98.9  &  65.6  \\
& {Ties-Merging} & {86.0 } &  76.5  &  85.0  &  89.4  &  95.9  &  90.3  &  83.3  &  99.0  &  68.8  \\
& {PCB-Merging} & \underline{87.5} &  \textbf{76.8}  &  86.2  &  89.4  &  96.5  &  88.3  &  91.0  &  98.6  &  \textbf{73.6}  \\
  & \textbf{{\method} (ours)} & \textbf{87.8}	& {75.4}	& 85.8	& \textbf{90.3}	& {96.6}	& \textbf{91.7}	& \textbf{91.2}	& \textbf{99.2}	& {72.4}  \\
  \bottomrule
\end{tabular}
\end{table*}

\subsection{Hyperparameter Settings}\label{app:hyperparameters}

\paragraph{Fully Fine-tuned NLP Model: T5 Merging}
{\method} is evaluated with a sparsity ratio $\eta=20\%$ and a scaling factor $\gamma=0.8$.
To reduce variance and ensure reliable evaluation, we report the experimental results of the average performance over different templates for each task, i.e., paws has 11 templates, qasc has 5 templates, quartz has 8 templates, story\_cloze has 5 templates, wiki\_qa has 5 templates, winogrande has 5 templates, and wsc has 10 templates.

\paragraph{PEFT of NLP Model: LORA Adapters Merging}
{\method} is evaluated with a sparsity ratio $\eta=30\%$ and a scaling factor $\gamma=0.5$.

\paragraph{Large Model: LLM Merging}
We follow the setup from \citet{du2024parameter} to merge these LLMs. For {\method}, we does not apply trimming step and set the scaling factor $\gamma=0.8$.

\paragraph{Fully Fine-tuned Vision Model: ViT Merging}
With sparse ratio $\eta=20\%$, we grid search over the scaling factor $\gamma$ in $[0.5, 1.2]$ with step size 0.1 and find the best performance with $\gamma=0.6$ for ViT-B/32 and $\gamma=0.7$ for ViT-L/14.

\end{document}